\documentclass{article}

\usepackage{xspace}

\usepackage{array, boldline, makecell}
\usepackage[svgnames, table]{xcolor}
\usepackage{multirow}
\usepackage{booktabs} 
\usepackage[table]{xcolor}
\usepackage{enumitem}
\usepackage{algorithm}
\usepackage{algpseudocode}  
\usepackage{wrapfig}
\usepackage{placeins}

\usepackage[utf8]{inputenc} 
\usepackage[T1]{fontenc}    
\usepackage{hyperref}       
\usepackage{url}            
\usepackage{amsfonts}       
\usepackage{nicefrac}       
\usepackage{microtype}      
\usepackage{xcolor}         
\usepackage{amsmath}
\usepackage{amssymb}
\usepackage{mathtools}
\usepackage{amsthm}
\usepackage{graphicx}
\usepackage{subcaption}
\usepackage{tabularx}
\usepackage{caption}

\makeatletter
\NewDocumentCommand{\supptitle}{s}{
\twocolumn[{
  \begin{center}
    \vspace*{-0.3cm}
    \rule{\textwidth}{0.05cm}\\[0.1cm]
    \textbf{- Appendix -}\\[0.2cm]
    {\Large \textbf{\mytitle}}\\[0.1cm]
    \rule{\textwidth}{0.05cm}\\[0.3cm]
  \end{center}
}]
}
\makeatother

\definecolor{LightCyan}{rgb}{0.88,1,1}
\definecolor{Blue}{rgb}{0, 0.5, 1}
\definecolor{Green}{rgb}{0.0, 0.8, 0.0 }
\definecolor{Red}{rgb}{0.95, 0.55, 0.6}
\definecolor{Skyblue}{rgb}{0.6, 0.6, 0.95 }
\definecolor{Beige}{rgb}{0.96, 0.96, 0.86}

\newcommand{\std}{0.7}

\newcommand{\system}{\texttt{Flux}}

\usepackage[normalem]{ulem}
\useunder{\uline}{\ul}{}

\usepackage{tcolorbox}
\tcbuselibrary{skins, breakable}

\usepackage[hang,flushmargin]{footmisc}

\usepackage{amsmath,amsfonts,bm}

\def\eqref#1{equation~\ref{#1}}

\def\1{\bm{1}}

\DeclareMathAlphabet{\mathsfit}{\encodingdefault}{\sfdefault}{m}{sl}
\SetMathAlphabet{\mathsfit}{bold}{\encodingdefault}{\sfdefault}{bx}{n}

\usepackage[preprint]{neurips_2026}

\title{Multimodal Federated Learning \\ under Dual-Axis Modality Missingness}

\author{
Adiba Orzikulova$^{*1}$\; Jaehyun Kwak$^{*1}$\,\\\textbf{ 
Jaemin Shin$^1$\; Yunqi Guo$^2$\; Xiaomin Ouyang$^3$\; Guoliang Xing$^2$}\\
 \textbf{   Steven Euijong Whang$^1$\; Sung-Ju Lee$^1$}\\
 $^1$KAIST, $^2$CUHK, $^3$HKUST \\
    \texttt{\{adiorz,jaehyun98,jaemin.shin,swhang,profsj\}@kaist.ac.kr}\\  
    \texttt{yunqiguo@cuhk.edu.hk}, \texttt{glxing@ie.cuhk.edu.hk}\\ 
    \texttt{xmouyang@cse.ust.hk}
}

\begin{document}

\maketitle

\begingroup
\renewcommand{\thefootnote}{$*$}
\footnotetext{Equal contribution}
\endgroup
\vspace{-2.0em}

\begin{abstract}
Multimodal federated learning~(FL) supports collaborative modeling in privacy-sensitive health-sensing and medical settings, but realistic deployments often exhibit \emph{dual-axis modality missingness}: clients have different modality sets, and individual samples may contain only subsets of the modalities available locally. Existing methods typically address these two axes separately. We propose \system{}, a multimodal federated learning framework built around two complementary components. First, \emph{modality-aware confidence tempering} learns sample-specific confidence for each modality through mask-aware unimodal supervision and fuses the confidence estimates from observed modalities into a sample-adaptive temperature that adjusts predictive sharpness according to evidence quality and completeness. Second, \emph{gradient-decoupled private adaptation} applies this temperature only to a client-private prediction pathway, while training the shared federated model with a standard, untempered objective. This enables sample-specific, client-local confidence adaptation without allowing confidence-dependent gradients to perturb shared representation learning. Across four multimodal datasets, \system{} achieves the highest average macro-F1 on every dataset, outperforming the strongest dataset-specific baseline by 0.8$\sim$2.2 points and by 1.6 points on average. Additional analyses demonstrate favorable calibration, temperature sensitivity to both modality missingness and input corruption, and more stable shared optimization under private-only tempering. Our code is available at \href{https://github.com/AdibaOrz/Flux}{https://github.com/AdibaOrz/Flux}.
\end{abstract}

\section{Introduction}\label{s:introduction}
Multimodal deep learning is increasingly deployed in privacy- and regulation-constrained settings, including medical and health-sensing applications~\citep{soenksen2022integrated, hemker2024healnet, shin2022mydj, yoon2025selfreplay, orzikulova2025bioq}. In these settings, data are often distributed across clients or silos such as hospitals, personal devices, or deployment sites, where centralizing raw data may be impractical or prohibited. Federated learning~(FL) therefore provides a practical framework for collaborative model training without exchanging raw data~\citep{mcmahan2017communication, kairouz2021advances}.

\begin{figure*}[t]
\centering
\includegraphics[width=.98\linewidth]{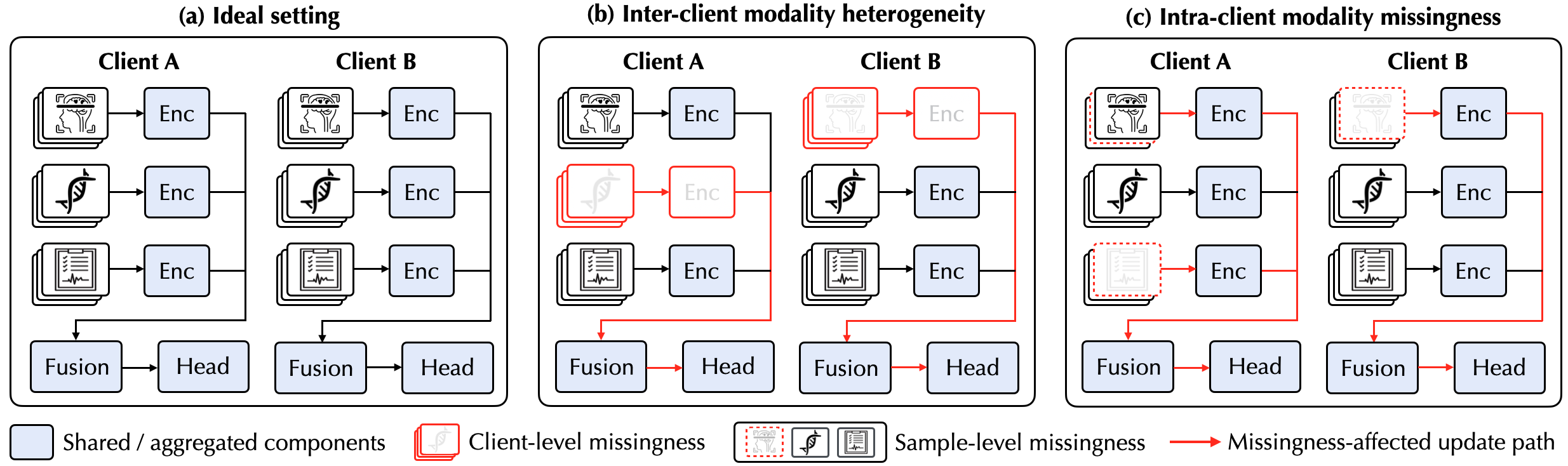}
\caption{
\emph{Dual-axis modality missingness} in multimodal FL:
(a) the complete-modality setting,
(b) \emph{inter-client modality heterogeneity}, and 
(c) \emph{intra-client modality missingness}.
Red arrows mark missingness-affected update paths.
}
\label{fig:dual_axis_missingness}
\vspace{-12pt}
\end{figure*}

\noindent A central challenge in realistic multimodal FL is modality missingness along two distinct but coupled axes. At the client level, clients may have access to different sets of modalities, a setting we refer to as \emph{inter-client modality heterogeneity}. For example, hospitals and devices may collect different subsets of imaging, genomic, clinical, or sensor modalities due to differences in acquisition protocols, infrastructure, or device capabilities. At the sample level, a particular instance may contain only a subset of the modalities available at its client, yielding \emph{intra-client modality missingness}. For example, a patient record may lack a modality because an examination was not performed, a test was cost-prohibitive, or data acquisition failed. These two forms of missingness often co-occur in realistic deployments~\citep{wang2017studentlife, vaizman2017recognizing}, creating a \emph{dual-axis modality missingness} regime. This regime violates the common simplifying assumption in multimodal FL that all clients share the same modality schema and that every sample contains complete observations for all modalities~\citep{xiong2022unified}. Figure~\ref{fig:dual_axis_missingness} contrasts this idealized setting~(a) with inter-client modality heterogeneity~(b) and intra-client modality missingness~(c).

Existing methods generally address these two axes separately. Approaches to inter-client modality heterogeneity coordinate learning across clients with different modality sets through modality-wise training or quality-aware aggregation~\citep{ouyang2023harmony, orzikulova2024federated}, but assume complete observations within each client. Conversely, methods for intra-client modality missingness improve robustness to partially observed samples, often through masking~\citep{bao2023multimodal, feng2023fedmultimodal}, while assuming a common modality schema across clients. These assumptions overlook their interaction during federated optimization: sample-level missingness determines which representations contribute to each local update, whereas client-level availability determines which modality branches and combinations each client can train. The server therefore aggregates shared components optimized under systematically different evidence distributions and modality coverage.

To address this coupled challenge, we propose \system{}, which separates sample-level evidence adaptation from shared federated optimization. First, \emph{modality-aware confidence tempering} learns task-relevant confidence for each observed modality through mask-aware unimodal supervision and combines these estimates into a sample-adaptive temperature. The temperature adjusts predictive sharpness when a sample contains fewer, noisier, or less informative modalities, thereby addressing intra-client variation in evidence completeness and quality.

Applying this confidence adjustment directly to the shared pathway, however, would introduce additional confidence-dependent variation into shared gradients. Because clients differ in both modality availability and sample-level missingness patterns, temperature-dependent changes in gradient magnitude and direction could amplify heterogeneity among the updates aggregated by the server. \emph{Gradient-decoupled private adaptation} prevents this coupling by applying the temperature only to a client-private prediction head operating on a detached fused representation, while training the shared encoders, fusion module, and global predictor with an untempered objective. Thus, confidence tempering guides evidence-aware private adaptation, whereas gradient decoupling prevents these sample-specific adjustments from distorting collaboratively learned representations.

We evaluate \system{} on four multimodal datasets spanning health-sensing and medical domains: PAMAP2~\citep{reiss2012introducing}, RealWorldHAR~\citep{sztyler2016body}, SleepEDF~\citep{goldberger2000physiobank, kemp2000analysis}, and ADNI~\citep{weiner2010alzheimer}, a heterogeneous biomedical dataset with naturally missing modalities. \system{} achieves the highest average macro-F1 on each dataset, outperforming the strongest dataset-specific baseline by 0.8$\sim$2.2 points, with an average gain of 1.6 points across datasets. Beyond predictive performance, our analyses show that \system{} achieves favorable calibration and learns a confidence-derived temperature that increases with modality incompleteness and input corruption. Private-only tempering also reduces shared-gradient instability and client-to-global update drift.

In summary, this work makes three contributions.
First, we formulate \emph{dual-axis modality missingness} in federated multimodal learning as a coupled regime in which client-level modality availability and sample-level modality observation jointly shape local optimization and server-side aggregation.
Second, we propose \system{}, which combines modality-aware confidence tempering with gradient-decoupled private adaptation, forming a sample-adaptive temperature from per-modality confidence while isolating confidence-tempered updates from shared parameters. Third, we demonstrate consistent performance gains across four health-sensing and medical datasets, including naturally incomplete biomedical data, and provide diagnostic analyses of calibration, sensitivity to evidence quality and shared-optimization stability.

\section{Problem Formulation}\label{s:problem_formulation}
We consider a federated multimodal learning setting with $K$ clients and a universal modality set $\mathcal{M}=\{1,\ldots,M\}$.
Each client $k\in \{1,\ldots,K\}$ holds a private dataset:
\[
\mathcal{D}_k=\{(\mathbf{x}_{k,i},y_{k,i},\mathbf{a}_{k,i})\}_{i=1}^{n_k}.
\]
For a sample $i$ on client $k$, $\mathbf{x}_{k,i}=(x_{k,i,1},\ldots,x_{k,i,M})$ is the multimodal input, $y_{k,i}$ is the label, and $\mathbf{a}_{k,i}\in\{0,1\}^{M}$ is the sample-level observation mask, where $a_{k,i,m}=1$ indicates that modality $m$ is observed for that sample. 
We further define a client-level availability mask
\(\mathbf{c}_k\in\{0,1\}^{M}\), where \(c_{k,m}=1\) indicates that modality $m$ is available at client $k$. A sample can contain only modalities available at its client:
\[
a_{k,i,m}\leq c_{k,m},
\;\;
\forall k\in\{1, \ldots, K\},i\in\{1,\ldots,n_k\},m\in \mathcal{M}.
\]
We assume that every sample contains at least one observed modality, i.e., $\sum_{m=1}^{M}a_{k,i,m}\geq1$.
Variation in \(\{\mathbf{c}_k\}_{k=1}^{K}\) characterizes
\emph{inter-client modality heterogeneity}. In particular, clients $k$ and $k'$ have different modality profiles when $\mathbf{c}_k\neq\mathbf{c}_{k'}$. 
By contrast, $c_{k,m}=1$ and $a_{k,i,m}=0$ indicate \emph{intra-client modality missingness}: a modality $m$ is available at client $k$ but absent from sample $i$.

The goal is to learn a federated predictor that performs well when  these two forms of missingness occur jointly. Let \(\theta\) denote the shared parameters communicated to and aggregated by the server, and let \(\phi_k\) denote the parameters maintained privately by client \(k\). For methods without client-specific components, \(\phi_k\) is omitted. The local empirical objective for client \(k\) is
\[
\mathcal{L}_k(\theta,\phi_k)
=
\frac{1}{n_k}
\sum_{i=1}^{n_k}
\ell\!\left(
f(\mathbf{x}_{k,i},\mathbf{a}_{k,i};\theta,\phi_k),
y_{k,i}
\right),
\]
where \(f(\cdot)\) is the multimodal predictor and
\(\ell(\cdot,\cdot)\) is the task loss. The corresponding federated objective is
\[
\min_{\theta,\{\phi_k\}_{k=1}^{K}}
\sum_{k=1}^{K}
\frac{n_k}{n}
\mathcal{L}_k(\theta,\phi_k),
\qquad
n=\sum_{k=1}^{K}n_k.
\]
\noindent At communication round \(t\), the server distributes the current shared parameters \(\theta^t\) to a selected client set \(\mathcal{S}^t\). Each selected client performs local optimization and returns its updated shared parameters \(\theta_k^{t+1}\) while retaining its private parameters locally. The server then computes
\[
\theta^{t+1}
=
\operatorname{Agg}\!\left(
\{\theta_k^{t+1}:k\in\mathcal{S}^t\}
\right),
\]
where \(\operatorname{Agg}(\cdot)\) denotes an FL aggregation rule, such as sample-size-weighted averaging~\citep{mcmahan2017communication}.

Under dual-axis modality missingness, the two masks affect optimization at different granularities. The sample-level mask \(\mathbf{a}_{k,i}\) determines which modality evidence contributes to each sample loss and the resulting gradient, whereas the client-level mask \(\mathbf{c}_k\) constrains which modality branches and cross-modal combinations client $k$ can optimize.
Consequently, the server aggregates shared components trained under client-dependent modality coverage and evidence distributions. This creates two coupled requirements: adapting each prediction to the evidence observed for that sample, and preventing client-specific modality availability and missingness from destabilizing shared federated optimization.
\system{} addresses them through modality-aware confidence tempering and gradient-decoupled private adaptation, respectively.

\section{\system{} Design}\label{s:method}
\begin{figure}[t!] 
\centering 
\includegraphics[width=0.6\textwidth]{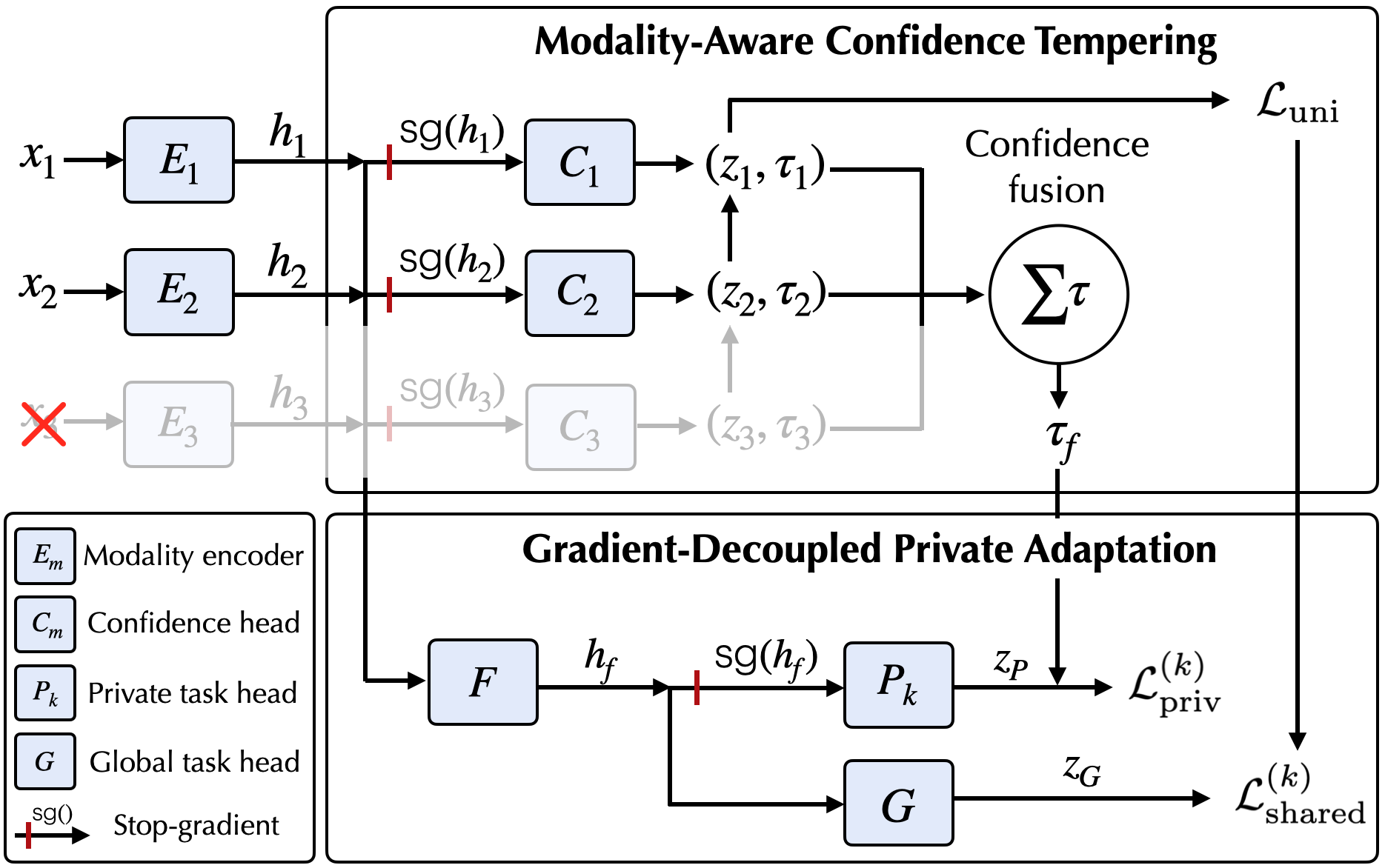} 
\caption{
Overview of \system{}. 
Faded paths denote missing modalities. Observed-modality temperatures are fused into $\tau_{f}$, 
which is applied to the private pathway operating on a detached fused representation.
}
\label{fig:method}
\vspace{-12pt}
\end{figure}

Figure~\ref{fig:method} summarizes \system{}. For each sample,  shared modality encoders transform the observed modalities into unimodal representations. \emph{Modality-aware confidence tempering} estimates confidence for each observed modality, and combines these estimates into a sample-adaptive temperature $\tau_f$ that reflects the completeness and reliability of the observed evidence. \emph{Gradient-decoupled private adaptation} then separates collaborative representation learning from confidence-dependent private adaptation: the shared multimodal pathway is trained with a standard, untempered task objective, whereas $\tau_f$ is applied only to the logits of a client-private task head operating on a detached fused representation.
The following subsections describe these components, followed by the federated training and inference procedures.
The training algorithm of \system{} is in Appendix~\ref{app:algorithm}.

\subsection{Modality-Aware Confidence Tempering}\label{ss:confidence_tempering}
Samples can vary not only in the number of observed modalities, but also in the reliability of their observed evidence. \system{} captures this variation by estimating an inverse-confidence temperature for each observed modality and combining the estimates into a sample-adaptive temperature. Unlike post-hoc sample-dependent temperature scaling~\citep{joy2023sample}, which predicts a single test-time temperature for the final logits of a fixed classifier, \system{} learns modality-level temperature during training and uses their fused value only to guide client-private adaptation.

We describe the method for a single training sample with label $y$ and observation mask
$\mathbf{a}\in\{0,1\}^{M}$, omitting client and sample indices for clarity. For each observed modality $m$, i.e., $a_m=1$, an encoder $E_m$ produces
\[
h_m=E_m(x_m).
\]

\paragraph{Per-modality confidence.}
A modality-specific auxiliary confidence head \(C_m\) receives the detached unimodal representation and outputs unimodal logits \(z_m\) and a scalar log-scale parameter \(\nu_m\):
\begin{equation}
\label{eq:confidence_head}
(z_m,\nu_m)
=
C_m\!\left(\operatorname{sg}(h_m)\right),
\end{equation}
where \(\operatorname{sg}(\cdot)\) denotes stop-gradient. 
Inspired by input-dependent uncertainty parametrization~\citep{kendall2017uncertainties}, we define the positive unimodal 
temperature
\[
\tau_m=\exp(\nu_m/2).
\]
We use $\tau_m$ as a deterministic inverse-confidence proxy: scaling the logits as $z_m / \tau_m$ sharpens the predictive distribution for lower $\tau_m$ and flattens it for higher $\tau_m$.

\paragraph{Mask-aware unimodal supervision.}
Let \(N_{\mathbf a}=\sum_{m=1}^{M}a_m\) denote the number of observed modalities. We train confidence heads of the observed modalities using a temperature-scaled unimodal classification loss:
\begin{equation}
\label{eq:unimodal_tempering}
\mathcal{L}_{\mathrm{uni}}
=
\frac{1}{N_{\mathbf a}}
\sum_{m=1}^{M}
a_m\,
\mathrm{CE}\!\left(
\frac{z_m}{\tau_m},
y
\right).
\end{equation}
Since \(h_m\) is detached, \(\mathcal{L}_{\mathrm{uni}}\) updates only $\{C_m\}$. This confines the auxiliary unimodal objective to the confidence heads, leaving the shared encoders optimized for multimodal fusion.

\paragraph{Fused multimodal temperature.}
Treating \(\tau_m^2\) as a modality-level relative uncertainty proxy, we combine the observed estimates using precision-inspired aggregation:
\begin{equation}
\label{eq:fused_temperature}
\tau_f
=
\left(
\sum_{m=1}^{M}
\frac{a_m}{\tau_m^2+\epsilon}
+\epsilon
\right)^{-1/2},
\end{equation}
where \(\epsilon>0\) ensures numerical stability. Lower-temperature modalities contribute more strongly to the aggregation, while fewer or less reliable observations yield a larger fused temperature. Thus, \(\tau_f\) reflects both the completeness
and estimated reliability of the observed evidence.

Applying \(\tau_f\) directly to the shared pathway would introduce additional confidence-dependent variation into shared gradients tied to client-specific observation patterns. This motivates the gradient separation introduced next.

\subsection{Gradient-Decoupled Private Adaptation}
\label{ss:private_adaptation}
\system{} separates shared task learning from confidence-tempered client-private adaptation during local optimization. The shared path is trained with untempered task logits, whereas \(\tau_f\) scales only the logits of a private head operating on a detached fused representation. This confines confidence-tempered gradients to the private head and prevents them from affecting parameters aggregated by the server.

\paragraph{Shared pathway.}
A mask-aware fusion module \(F\) combines only the observed unimodal representations:
\[
h_f
=
F\!\left(
\{(a_m,h_m)\}_{m=1}^{M}
\right).
\]
The shared task head \(G\) then produces
\[
z_G=G(h_f).
\]
On client \(k\), the objective to update shared parameters is
\begin{equation}
\label{eq:shared_loss}
\mathcal{L}_{\mathrm{shared}}^{(k)}
=
\mathbb{E}_{(\mathbf{x},y,\mathbf{a})\sim\mathcal{D}_k}
\left[
\mathrm{CE}(z_G,y)
+
\mathcal{L}_{\mathrm{uni}}
\right].
\end{equation}
The untempered task loss \(\mathrm{CE}(z_G,y)\) updates the shared encoders, fusion module, and task head. Because \(h_m\) is detached in \(\mathcal{L}_{\mathrm{uni}}\), the auxiliary loss updates only the confidence heads and does not push the shared modality encoders toward unimodal prediction.

\paragraph{Private pathway.}
Each client \(k\) maintains a private task head \(P_k\) with parameters \(\phi_k\). The head operates on the detached fused representation:
\begin{equation}
\label{eq:private_logits}
z_{P,k}
=
P_k\!\left(
\operatorname{sg}(h_f)
\right).
\end{equation}
The private objective scales these logits using the detached fused temperature:
\begin{equation}
\label{eq:private_loss}
\mathcal{L}_{\mathrm{priv}}^{(k)}
=
\mathbb{E}_{(\mathbf{x},y,\mathbf{a})\sim\mathcal{D}_k}
\left[
\mathrm{CE}\!\left(
\frac{z_{P,k}}
{\operatorname{sg}(\tau_f)},
y
\right)
\right].
\end{equation}
Thus, $\tau_f$ modulates the private-head training according to the estimated completeness and reliability of the observed evidence, while $\mathcal{L}_\mathrm{priv}^{(k)}$ updates only $\phi_k$. Detaching \(h_f\) blocks gradients to the modality encoders and fusion module, while detaching \(\tau_f\) blocks gradients to the confidence heads.

\paragraph{Rationale for private-only tempering.}
If \(\tau_f\) were applied to the shared logits, then, treating it as
constant, the gradient with respect to \(z_G\) would be
\[
\frac{\partial}{\partial z_G}
\mathrm{CE}\!\left(
\frac{z_G}{\tau_f},
y
\right)
=
\frac{1}{\tau_f}
\left(
p_\tau-e_y
\right),
\]
where
\(
p_\tau
=
\operatorname{softmax}(z_G/\tau_f)
\) and \(e_y\) is the one-hot label vector. With \(p=\operatorname{softmax}(z_G)\), this can be written as: 
\[
\frac{1}{\tau_f}(p_\tau-e_y)
=
\frac{1}{\tau_f}(p-e_y)
+
\frac{1}{\tau_f}(p_\tau-p).
\]
Relative to the untempered gradient \(p-e_y\), \(\tau_f\) both rescales the gradient and introduces a temperature-dependent correction. In multiclass settings, this correction is generally not collinear with the original gradient, so tempering can alter both its magnitude and direction. Because \(\tau_f\) varies with each sample's observed evidence, applying it to the shared pathway would introduce additional sample-dependent variation into the updates aggregated across clients.
\system{} therefore confines \(\tau_f\) to the private-head loss, preventing temperature-dependent modification of shared task gradients.

\subsection{Federated Training and Inference}
\label{ss:optimization}
\paragraph{Local training.}
At communication round \(t\), each selected client \(k\) sets its local shared parameters to \(\theta^{(t)}\) and loads its retained private parameters \(\phi_k\). For each minibatch, the client computes the unimodal representations, modality-level confidence outputs, fused representation \(h_f\), fused temperature \(\tau_f\), and shared and private logits. It updates the shared parameters using \(\mathcal{L}_{\mathrm{shared}}^{(k)}\) and the private parameters
\(\phi_k\) using \(\mathcal{L}_{\mathrm{priv}}^{(k)}\). After local optimization, client \(k\) returns its updated shared parameters \(\theta_k^{(t+1)}\) to the server and retains the updated \(\phi_k\) locally for the next round.

\paragraph{Server aggregation.}
Let
\[
n_{k,m}
=
\sum_{i=1}^{n_k} a_{k,i,m}
\]
denote the number of samples in client $k$'s local training set for which modality $m$ is observed, and define
\[
\mathcal{S}_m^t
=
\{k\in\mathcal{S}^t:n_{k,m}>0\}
\]
as the selected clients that contribute an update for modality \(m\).
For each modality-specific parameter block
\[
\theta_m=
\left(
{\theta}_{E_m},
{\theta}_{C_m}
\right),
\]
containing the parameters of encoder \(E_m\) and confidence head
\(C_m\), the server computes:
\[
\theta_m^{(t+1)}
=
\begin{cases}
\displaystyle
\sum_{k\in\mathcal{S}_m^t}
\frac{n_{k,m}}
{\sum_{j\in\mathcal{S}_m^t}n_{j,m}}
\theta_{k,m}^{(t+1)},
&
\mathcal{S}_m^t\neq\emptyset,
\\[8pt]
\theta_m^{(t)},
&
\mathcal{S}_m^t=\emptyset.
\end{cases}
\]
The remaining shared parameter blocks, namely the fusion module and shared task head, are aggregated using client-level sample weights:
\[
\theta_{F,G}^{(t+1)}
=
\sum_{k\in\mathcal{S}^t}
\frac{n_k}
{\sum_{j\in\mathcal{S}^t}n_j}
\theta_{k,F,G}^{(t+1)},
\]
where \(\theta_{F,G}\) contains the parameters of \(F\) and \(G\).
The private parameters \(\{\phi_k\}\) are never communicated or aggregated.

\paragraph{Inference.}
At test time, client \(k\) combines the shared and private logits to obtain the predictive distribution:
\[
\begin{aligned}
\hat z_k &= z_G + z_{P,k},
&\quad
\hat p_k &= \operatorname{softmax}(\hat z_k),\hat y_k &= \arg\max_c [\hat p_k]_c.
\end{aligned}
\]
We do not train the model using a joint loss on \(z_G+z_{P,k}\). Such a loss would make the gradient of the shared head depend on the client-private logits, coupling updates to the shared pathway with each client's private model state. The confidence heads and \(\tau_f\) are used only to guide private-head optimization during training and are not required at inference. Confidence estimation is therefore learned collaboratively, while confidence-tempered adaptation remains client-local. Relative to the shared model, inference requires only the lightweight private task head.

\section{Experiments}\label{s:experiments}

\begin{table*}[t!]
\caption{
Macro-F1 under dual-axis missingness on three health-sensing datasets. \textbf{H} and \textbf{I} denote inter-client heterogeneity levels \{Homogeneous, Moderate, High\} and intra-client incompleteness levels \{Moderate, High\}.
}
\label{tab:main_results}
\centering

\renewcommand{\arraystretch}{1.15}
\setlength{\tabcolsep}{2.2pt}
\fontsize{8pt}{10pt}\selectfont

\begin{subtable}{\textwidth}
\centering

\begin{tabular*}{.98\textwidth}{@{\extracolsep{\fill}}lcccccccc@{}}
\toprule
\multicolumn{9}{c}{\textbf{PAMAP2}} \\
\midrule
\textbf{(H, I)} &
\textbf{FedAvg} & \textbf{FedProx} & \textbf{MOON} &
\textbf{FedPer} & \textbf{FedRoD} &
\textbf{PmcmFL} & \textbf{PEPSY} &
\textbf{Flux (Ours)} \\
\midrule
(Hom, Mod)   &
0.722 \scalebox{\std}{$\pm$ 0.007} & 0.722 \scalebox{\std}{$\pm$ 0.003} & 0.723 \scalebox{\std}{$\pm$ 0.011} &
0.694 \scalebox{\std}{$\pm$ 0.012} & 0.749 \scalebox{\std}{$\pm$ 0.009} &
0.723 \scalebox{\std}{$\pm$ 0.008} &
 0.722 \scalebox{\std}{$\pm$ 0.015} &
\textbf{0.761 \scalebox{\std}{$\pm$ 0.006}} \\
(Hom, High)  &
0.596 \scalebox{\std}{$\pm$ 0.025} & 0.600 \scalebox{\std}{$\pm$ 0.022} & 0.610 \scalebox{\std}{$\pm$ 0.012} &
0.453 \scalebox{\std}{$\pm$ 0.011} & 0.609 \scalebox{\std}{$\pm$ 0.012} &
0.605 \scalebox{\std}{$\pm$ 0.007} &
 0.590 \scalebox{\std}{$\pm$ 0.013} &
\textbf{0.641 \scalebox{\std}{$\pm$ 0.010}} \\
(Mod, Mod)   &
0.634 \scalebox{\std}{$\pm$ 0.008} & 0.636 \scalebox{\std}{$\pm$ 0.004} & 0.636 \scalebox{\std}{$\pm$ 0.014} &
0.644 \scalebox{\std}{$\pm$ 0.013} & 0.664 \scalebox{\std}{$\pm$ 0.012} &
0.643 \scalebox{\std}{$\pm$ 0.015} &
0.641 \scalebox{\std}{$\pm$ 0.014} &
\textbf{0.683 \scalebox{\std}{$\pm$ 0.012}} \\
(Mod, High)  &
0.574 \scalebox{\std}{$\pm$ 0.008} & 0.569 \scalebox{\std}{$\pm$ 0.013} & 0.561 \scalebox{\std}{$\pm$ 0.018} &
0.490 \scalebox{\std}{$\pm$ 0.006} & 0.588 \scalebox{\std}{$\pm$ 0.013} &
0.578 \scalebox{\std}{$\pm$ 0.020} &
 0.539 \scalebox{\std}{$\pm$ 0.009} &
\textbf{0.596 \scalebox{\std}{$\pm$ 0.024}} \\
(High, Mod)  &
0.621 \scalebox{\std}{$\pm$ 0.008} & 0.620 \scalebox{\std}{$\pm$ 0.011} & 0.616 \scalebox{\std}{$\pm$ 0.007} &
0.614 \scalebox{\std}{$\pm$ 0.011} & 0.636 \scalebox{\std}{$\pm$ 0.011} &
0.618 \scalebox{\std}{$\pm$ 0.006} &
 0.627 \scalebox{\std}{$\pm$ 0.017} &
\textbf{0.655 \scalebox{\std}{$\pm$ 0.006}} \\
(High, High) &
0.499 \scalebox{\std}{$\pm$ 0.014} & 0.497 \scalebox{\std}{$\pm$ 0.016} & 0.494 \scalebox{\std}{$\pm$ 0.014} &
0.440 \scalebox{\std}{$\pm$ 0.009} & 0.500 \scalebox{\std}{$\pm$ 0.017} &
0.514 \scalebox{\std}{$\pm$ 0.019} &
 0.508 \scalebox{\std}{$\pm$ 0.011} &
\textbf{0.520 \scalebox{\std}{$\pm$ 0.014}} \\
\midrule
\textbf{Average} &
0.608 \scalebox{\std}{$\pm$ 0.011} & 0.607 \scalebox{\std}{$\pm$ 0.011} & 0.607 \scalebox{\std}{$\pm$ 0.013} &
0.556 \scalebox{\std}{$\pm$ 0.010} & 0.624 \scalebox{\std}{$\pm$ 0.012} &
0.614 \scalebox{\std}{$\pm$ 0.012} &
 0.604 \scalebox{\std}{$\pm$ 0.013} &
\textbf{0.642 \scalebox{\std}{$\pm$ 0.012}} \\
\bottomrule
\end{tabular*}
\end{subtable}

\vspace{0.6em}

\begin{subtable}{\textwidth}
\centering
\begin{tabular*}{.98\textwidth}{@{\extracolsep{\fill}}lcccccccc@{}}
\toprule
\multicolumn{9}{c}{\textbf{SleepEDF}} \\
\midrule
\textbf{(H, I)} &
\textbf{FedAvg} & \textbf{FedProx} & \textbf{MOON} &
\textbf{FedPer} & \textbf{FedRoD} &
\textbf{PmcmFL} & \textbf{PEPSY} &
\textbf{Flux (Ours)} \\
\midrule
(Hom, Mod)   &
0.589 \scalebox{\std}{$\pm$ 0.010} & 0.589 \scalebox{\std}{$\pm$ 0.007} & 0.594 \scalebox{\std}{$\pm$ 0.008} &
0.580 \scalebox{\std}{$\pm$ 0.011} & 0.600 \scalebox{\std}{$\pm$ 0.009} &
0.601 \scalebox{\std}{$\pm$ 0.006} &
0.589 \scalebox{\std}{$\pm$ 0.006} &
\textbf{0.616 \scalebox{\std}{$\pm$ 0.008}} \\
(Hom, High)  &
0.526 \scalebox{\std}{$\pm$ 0.007} & 0.534 \scalebox{\std}{$\pm$ 0.007} & 0.527 \scalebox{\std}{$\pm$ 0.007} &
0.510 \scalebox{\std}{$\pm$ 0.004} & 0.523 \scalebox{\std}{$\pm$ 0.012} &
0.547 \scalebox{\std}{$\pm$ 0.005} &
0.537 \scalebox{\std}{$\pm$ 0.015} &
\textbf{0.557 \scalebox{\std}{$\pm$ 0.007}} \\
(Mod, Mod)   &
0.537 \scalebox{\std}{$\pm$ 0.007} & 0.531 \scalebox{\std}{$\pm$ 0.008} & 0.537 \scalebox{\std}{$\pm$ 0.003} &
0.520 \scalebox{\std}{$\pm$ 0.006} & 0.528 \scalebox{\std}{$\pm$ 0.005} &
0.528 \scalebox{\std}{$\pm$ 0.004} &
0.513 \scalebox{\std}{$\pm$ 0.012} &
\textbf{0.552 \scalebox{\std}{$\pm$ 0.009}} \\
(Mod, High)  &
0.451 \scalebox{\std}{$\pm$ 0.013} & 0.438 \scalebox{\std}{$\pm$ 0.008} & 0.448 \scalebox{\std}{$\pm$ 0.010} &
0.434 \scalebox{\std}{$\pm$ 0.009} & 0.439 \scalebox{\std}{$\pm$ 0.008} &
0.439 \scalebox{\std}{$\pm$ 0.010} &
0.442 \scalebox{\std}{$\pm$ 0.009} &
\textbf{0.467 \scalebox{\std}{$\pm$ 0.007}} \\
(High, Mod)  &
0.519 \scalebox{\std}{$\pm$ 0.008} & 0.512 \scalebox{\std}{$\pm$ 0.008} & 0.514 \scalebox{\std}{$\pm$ 0.009} &
0.518 \scalebox{\std}{$\pm$ 0.005} & 0.535 \scalebox{\std}{$\pm$ 0.010} &
0.489 \scalebox{\std}{$\pm$ 0.004} &
0.482 \scalebox{\std}{$\pm$ 0.008} &
\textbf{0.540 \scalebox{\std}{$\pm$ 0.008}} \\
(High, High) &
0.498 \scalebox{\std}{$\pm$ 0.014} & 0.497 \scalebox{\std}{$\pm$ 0.010} & 0.495 \scalebox{\std}{$\pm$ 0.012} &
0.463 \scalebox{\std}{$\pm$ 0.008} & 0.494 \scalebox{\std}{$\pm$ 0.013} &
0.483 \scalebox{\std}{$\pm$ 0.005} &
0.478 \scalebox{\std}{$\pm$ 0.016} &
\textbf{0.518 \scalebox{\std}{$\pm$ 0.003}} \\
\midrule
\textbf{Average} &
0.520 \scalebox{\std}{$\pm$ 0.010} & 0.517 \scalebox{\std}{$\pm$ 0.008} & 0.519 \scalebox{\std}{$\pm$ 0.008} &
0.504 \scalebox{\std}{$\pm$ 0.007} & 0.520 \scalebox{\std}{$\pm$ 0.009} &
0.515 \scalebox{\std}{$\pm$ 0.006} &
0.507 \scalebox{\std}{$\pm$ 0.011} &
\textbf{0.542 \scalebox{\std}{$\pm$ 0.007}} \\
\bottomrule
\end{tabular*}
\end{subtable}

\vspace{0.6em}

\begin{subtable}{\textwidth}
\centering
\begin{tabular*}{.98\textwidth}{@{\extracolsep{\fill}}lcccccccc@{}}
\toprule
\multicolumn{9}{c}{\textbf{RealWorldHAR}} \\
\midrule
\textbf{(H, I)} &
\textbf{FedAvg} & \textbf{FedProx} & \textbf{MOON} &
\textbf{FedPer} & \textbf{FedRoD} &
\textbf{PmcmFL} & \textbf{PEPSY} &
\textbf{Flux (Ours)} \\
\midrule
(Hom, Mod)   &
0.838 \scalebox{\std}{$\pm$ 0.002} & 0.832 \scalebox{\std}{$\pm$ 0.010} & 0.836 \scalebox{\std}{$\pm$ 0.014} &
0.852 \scalebox{\std}{$\pm$ 0.016} & \textbf{0.885 \scalebox{\std}{$\pm$ 0.003}} &
0.840 \scalebox{\std}{$\pm$ 0.006} &
0.839 \scalebox{\std}{$\pm$ 0.006} &
0.867 \scalebox{\std}{$\pm$ 0.011} \\
(Hom, High)  &
0.686 \scalebox{\std}{$\pm$ 0.007} & 0.673 \scalebox{\std}{$\pm$ 0.004} & 0.670 \scalebox{\std}{$\pm$ 0.005} &
0.528 \scalebox{\std}{$\pm$ 0.015} & 0.704 \scalebox{\std}{$\pm$ 0.023} &
0.702 \scalebox{\std}{$\pm$ 0.013} &
\textbf{0.772 \scalebox{\std}{$\pm$ 0.009}} &
0.740 \scalebox{\std}{$\pm$ 0.004} \\
(Mod, Mod)   &
0.844 \scalebox{\std}{$\pm$ 0.006} & 0.847 \scalebox{\std}{$\pm$ 0.001} & 0.844 \scalebox{\std}{$\pm$ 0.004} &
0.820 \scalebox{\std}{$\pm$ 0.015} & 0.865 \scalebox{\std}{$\pm$ 0.011} &
0.842 \scalebox{\std}{$\pm$ 0.006} &
0.853 \scalebox{\std}{$\pm$ 0.005} &
\textbf{0.858 \scalebox{\std}{$\pm$ 0.007}} \\
(Mod, High)  &
0.674 \scalebox{\std}{$\pm$ 0.008} & 0.673 \scalebox{\std}{$\pm$ 0.012} & 0.669 \scalebox{\std}{$\pm$ 0.008} &
0.616 \scalebox{\std}{$\pm$ 0.011} & 0.693 \scalebox{\std}{$\pm$ 0.004} &
0.666 \scalebox{\std}{$\pm$ 0.005} &
\textbf{0.760 \scalebox{\std}{$\pm$ 0.017}} &
0.709 \scalebox{\std}{$\pm$ 0.006} \\
(High, Mod)  &
0.838 \scalebox{\std}{$\pm$ 0.008} & 0.840 \scalebox{\std}{$\pm$ 0.003} & 0.843 \scalebox{\std}{$\pm$ 0.006} &
0.835 \scalebox{\std}{$\pm$ 0.002} & 0.851 \scalebox{\std}{$\pm$ 0.006} &
0.834 \scalebox{\std}{$\pm$ 0.006} &
0.803 \scalebox{\std}{$\pm$ 0.005} &
\textbf{0.855 \scalebox{\std}{$\pm$ 0.007}} \\
(High, High) &
0.709 \scalebox{\std}{$\pm$ 0.007} & 0.716 \scalebox{\std}{$\pm$ 0.010} & 0.709 \scalebox{\std}{$\pm$ 0.009} &
0.720 \scalebox{\std}{$\pm$ 0.012} & 0.747 \scalebox{\std}{$\pm$ 0.009} &
0.722 \scalebox{\std}{$\pm$ 0.013} &
0.710 \scalebox{\std}{$\pm$ 0.008} &
\textbf{0.766 \scalebox{\std}{$\pm$ 0.004}} \\
\midrule
\textbf{Average} &
0.765 \scalebox{\std}{$\pm$ 0.006} & 0.763 \scalebox{\std}{$\pm$ 0.007} & 0.762 \scalebox{\std}{$\pm$ 0.008} &
0.729 \scalebox{\std}{$\pm$ 0.012} & 0.791 \scalebox{\std}{$\pm$ 0.009} &
0.768 \scalebox{\std}{$\pm$ 0.008} &
0.789 \scalebox{\std}{$\pm$ 0.008} &
\textbf{0.799 \scalebox{\std}{$\pm$ 0.006}} \\
\bottomrule
\end{tabular*}
\end{subtable}
\vspace{-5pt}
\normalsize
\end{table*}
\begin{table*}[t!]
\caption{
Macro-F1 on ADNI with natural intra-client incompleteness under varying inter-client heterogeneity. Extreme assigns each client a unique modality combination.
}
\label{tab:adni_results}
\centering

\renewcommand{\arraystretch}{1.12}
\setlength{\tabcolsep}{2.2pt}
\fontsize{8pt}{10pt}\selectfont

\begin{tabular*}{.98\textwidth}{@{\extracolsep{\fill}}lcccccccc@{}}
\toprule
\textbf{H} &
\textbf{FedAvg} &
\textbf{FedProx} &
\textbf{MOON} &
\textbf{FedPer} &
\textbf{FedRoD} &
\textbf{PmcmFL} &
\textbf{PEPSY} &
\textbf{Flux (Ours)} \\
\midrule
Homogen. &
0.568 \scalebox{\std}{$\pm$ 0.022} &
0.570 \scalebox{\std}{$\pm$ 0.014} &
\textbf{0.572 \scalebox{\std}{$\pm$ 0.017}} &
0.511 \scalebox{\std}{$\pm$ 0.034} &
0.542 \scalebox{\std}{$\pm$ 0.019} &
0.428 \scalebox{\std}{$\pm$ 0.080} &
0.509 \scalebox{\std}{$\pm$ 0.001} &
0.568 \scalebox{\std}{$\pm$ 0.012} \\
Moderate &
0.511 \scalebox{\std}{$\pm$ 0.031} &
0.527 \scalebox{\std}{$\pm$ 0.020} &
0.532 \scalebox{\std}{$\pm$ 0.012} &
0.505 \scalebox{\std}{$\pm$ 0.018} &
0.521 \scalebox{\std}{$\pm$ 0.022} &
0.279 \scalebox{\std}{$\pm$ 0.085} &
0.492 \scalebox{\std}{$\pm$ 0.003} &
\textbf{0.533 \scalebox{\std}{$\pm$ 0.013}} \\
High &
0.516 \scalebox{\std}{$\pm$ 0.037} &
0.529 \scalebox{\std}{$\pm$ 0.010} &
0.530 \scalebox{\std}{$\pm$ 0.014} &
0.469 \scalebox{\std}{$\pm$ 0.026} &
0.498 \scalebox{\std}{$\pm$ 0.009} &
0.203 \scalebox{\std}{$\pm$ 0.006} &
0.419 \scalebox{\std}{$\pm$ 0.050} &
\textbf{0.533 \scalebox{\std}{$\pm$ 0.020}} \\
Extreme &
0.330 \scalebox{\std}{$\pm$ 0.029} &
0.350 \scalebox{\std}{$\pm$ 0.027} &
0.330 \scalebox{\std}{$\pm$ 0.024} &
0.379 \scalebox{\std}{$\pm$ 0.039} &
0.389 \scalebox{\std}{$\pm$ 0.025} &
0.223 \scalebox{\std}{$\pm$ 0.048} &
0.293 \scalebox{\std}{$\pm$ 0.043} &
\textbf{0.395 \scalebox{\std}{$\pm$ 0.032}} \\
\midrule
\textbf{Average} &
0.481 \scalebox{\std}{$\pm$ 0.030} &
0.494 \scalebox{\std}{$\pm$ 0.018} &
0.491 \scalebox{\std}{$\pm$ 0.017} &
0.466 \scalebox{\std}{$\pm$ 0.029} &
0.487 \scalebox{\std}{$\pm$ 0.019} &
0.283 \scalebox{\std}{$\pm$ 0.055} &
0.428 \scalebox{\std}{$\pm$ 0.024} &
\textbf{0.508 \scalebox{\std}{$\pm$ 0.019}} \\
\bottomrule
\end{tabular*}

\vspace{-15pt}
\normalsize
\end{table*}

\paragraph{Datasets and baselines.} 
We evaluate \system{} across four multimodal datasets spanning health-sensing and medical domains: PAMAP2~\citep{reiss2012introducing}, RealWorldHAR~\citep{sztyler2016body}, SleepEDF~\citep{goldberger2000physiobank, kemp2000analysis}, and data obtained from Alzheimer's Disease NeuroImaging Initiative~(ADNI) database~\citep{weiner2010alzheimer}, which contains naturally incomplete multimodal records.
For the three health-sensing datasets, we construct controlled dual-axis missingness regimes.
For ADNI, we retain the naturally occurring sample-level missingness and introduce inter-client heterogeneity by removing selected modalities from all samples at each client.
We benchmark \system{} against baselines from four methodological families: the canonical FL method FedAvg~\citep{mcmahan2017communication}; non-IID–robust methods FedProx~\citep{li2020federatedprox} and MOON~\citep{li2021model}; personalization and decoupling approaches FedPer~\citep{arivazhagan2019federated} and FedRoD~\citep{chen2022on}; and missing-modality FL methods PmcmFL~\citep{bao2023multimodal} and PEPSY~\citep{nguyenlearning}. 
Details of the dual-axis missingness construction are provided in Appendix~\ref{app:missingness_simulation}, while additional information on the datasets, baselines, training, and evaluation protocols is provided in Appendix~\ref{app:experiment_details}.

\paragraph{Models and learning.}
For health-sensing datasets, we use 1D CNN encoders, a masked attention-based fusion module~\citep{bahdanau2014neural}, and a two-layer MLP classifier. FedRoD and \system{} add a client-private task head to the architecture, while \system{} additionally includes modality-specific confidence heads during training. We train for 200 global rounds, sampling 30$\sim$50\% of clients per round. Each selected client performs three local epochs using SGD.
ADNI contains four heterogeneous modalities: MRI, genomic profiles, clinical assessments, and biospecimen measurements. We therefore use modality-specific encoders for these heterogeneous inputs, following prior work~\citep{yun2024flex}. We train ADNI models for 150 global rounds with full client participation and one local epoch per client using Adam.
Experiments were conducted using Intel Xeon CPUs and NVIDIA RTX 3090 GPUs.

\section{Results}\label{s:results}
\paragraph{Overall Results.}
As shown in Table~\ref{tab:main_results}, \system{} achieves the highest average macro-F1 across the six dual-axis regimes on each health-sensing dataset: $0.642$ on PAMAP2, $0.542$ on SleepEDF, and $0.799$ on RealWorldHAR. It outperforms all baselines in every regime on PAMAP2 and SleepEDF, and leads overall on RealWorldHAR, including both high inter-client-heterogeneity settings.
Table~\ref{tab:adni_results} reports results on ADNI, where natural sample-level missingness is retained while client-level modality availability is varied. \system{} achieves the highest average macro-F1 of $0.508$, remains competitive in the homogeneous setting, and performs best under moderate, high, and extreme inter-client heterogeneity. In the extreme setting, where each client has a unique modality composition, \system{} reaches $0.395$, compared with $0.389$ for FedRoD. Overall,
these results demonstrate consistent performance across varying combinations of sample-level incompleteness and client-level modality heterogeneity.

\vspace{-3pt}
\paragraph{Confidence and Calibration Diagnostics.}

\begin{figure*}[t]
    \centering

    \begin{minipage}[t]{0.48\textwidth}
        \centering
        \includegraphics[width=\linewidth]
        {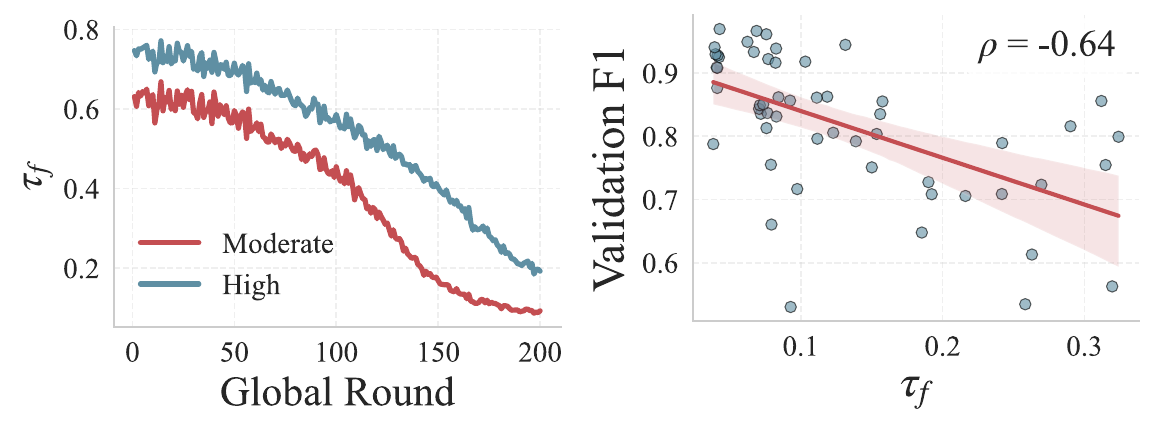}
        \caption{
        Fused temperature $\tau_f$ versus intra-client incompleteness
        (left) and client-level validation macro-F1 (right).
        }
        \label{fig:temperature_missing_perf_corr}
    \end{minipage}
    \hfill
    \begin{minipage}[t]{0.48\textwidth}
        \centering
        \includegraphics[width=\linewidth]
        {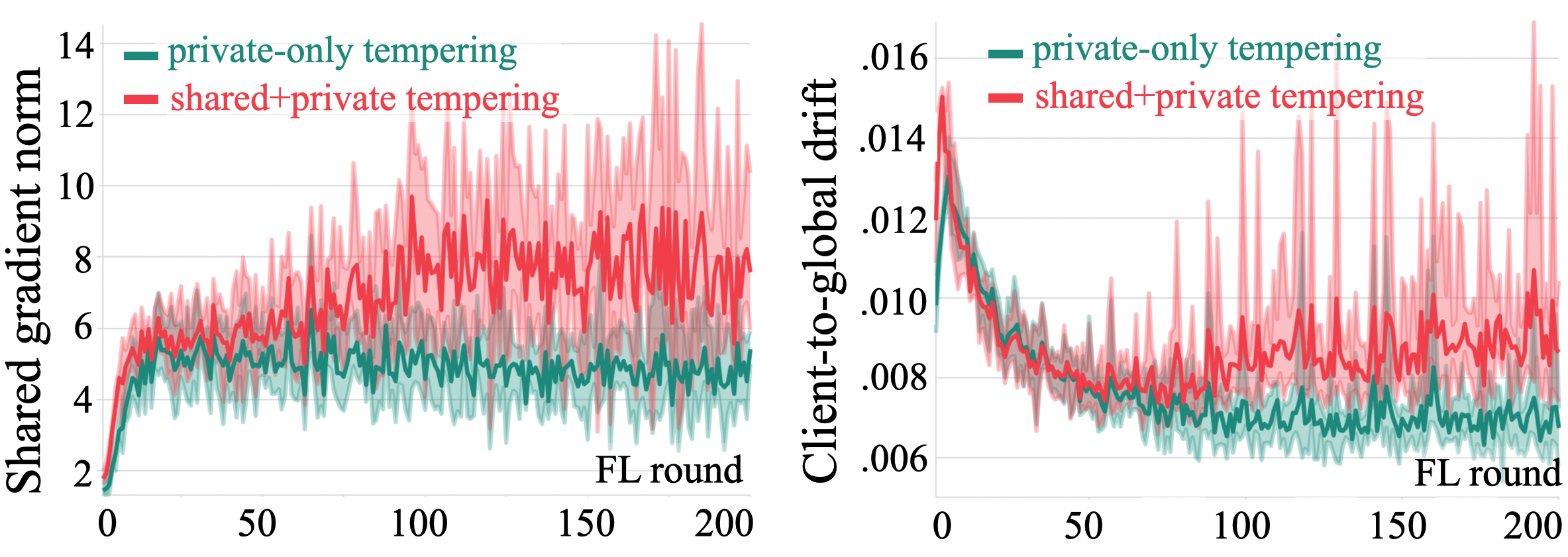}
        \caption{
        Shared-gradient norm (left) and client-to-global update drift
        (right) on the PAMAP2 dataset.
        }
        \label{fig:optimization_behavior}
    \end{minipage}
    \vspace{-7pt}
\end{figure*}

Figure~\ref{fig:temperature_missing_perf_corr} examines whether the learned temperature tracks variation in modality completeness and predictive performance. Under moderate inter-client heterogeneity, higher intra-client
incompleteness is consistently associated with larger $\tau_f$, indicating lower confidence when fewer modalities are observed. Across all six dual-axis regimes, client-level $\tau_f$ is also negatively correlated with validation F1 (Spearman $\rho=-0.64$, $p<0.001$). Thus, $\tau_f$ systematically reflects both modality incompleteness and predictive performance.

We also assess the calibration of the final combined predictions. Although \(\tau_f\) is used only during training, it shapes private-head optimization and may affect the combined shared and private logits at inference. We evaluate negative log-likelihood~(NLL), expected calibration error~(ECE), and Brier score on these logits. Table~\ref{tab:calibration} shows favorable calibration performance: \system{} achieves the best ECE (\(0.162\)) and Brier score (\(0.487\)), and the second-best NLL (\(1.379\)).

\paragraph{Effect of Tempering Placement on Shared Optimization.}
\label{ss:optimization_analysis}
To isolate the effect of tempering placement, we compare \emph{private-only tempering} (\system{}) with \emph{shared+private tempering}, which also applies \(\tau_f\) to the shared task head. Both variants use the same architecture, parameter partitioning, aggregation rule, and hyperparameters on PAMAP2 under the (High, High) dual-axis missingness regime. We measure (i) the mean \(\ell_2\) norm of gradients in the fusion module and shared task head and (ii) the mean relative client-to-global update drift over all shared parameters. Figure~\ref{fig:optimization_behavior} shows that shared+private tempering yields larger and more variable gradient norms and greater update drift, whereas private-only tempering remains lower and more stable after the initial rounds. These results indicate that applying \(\tau_f\) to both the shared and private heads increases update heterogeneity, while restricting it to private-head training yields a more stable shared optimization trajectory.


\begin{table*}[t!]
    \centering

    \begin{minipage}[t]{0.49\textwidth}
        \centering
        \caption{
        Calibration performance on PAMAP2 dataset under (Mod, Mod)
        missingness, averaged over five random seeds.
        }
        \label{tab:calibration}
        \vspace{-5pt}

        \setlength{\tabcolsep}{2.8pt}
        \renewcommand{\arraystretch}{1.18}
        \small
        \begin{tabular}{lccccc}
            \toprule
            Metric & FedAvg & FedRoD & PmcmFL & PEPSY & \system{} \\
            \midrule
            NLL $\downarrow$
            & 1.612
            & \textbf{1.366}
            & 1.447
            & 2.165
            & 1.379 \\
            ECE $\downarrow$
            & 0.193
            & 0.163
            & 0.177
            & 0.220
            & \textbf{0.162} \\
            Brier $\downarrow$
            & 0.525
            & 0.500
            & 0.514
            & 0.533
            & \textbf{0.487} \\
            \bottomrule
        \end{tabular}
    \end{minipage}
    \hfill
    \begin{minipage}[t]{0.49\textwidth}
        \centering
        \caption{
        Mean fused temperature $\tau_f$ under Gaussian input corruption
        on PAMAP2 with all modalities observed.
        }
        \label{tab:noisy_modalities}
        \vspace{-8pt}

        \setlength{\tabcolsep}{6pt}
        \small
        \begin{tabular}{@{}lccc@{}}
            \toprule
            FL Round
            & $\boldsymbol{\alpha=0.0}$
            & $\boldsymbol{\alpha=0.5}$
            & $\boldsymbol{\alpha=0.9}$ \\
            \midrule
            50  & 0.300 & 0.309 & 0.312 \\
            100 & 0.063 & 0.075 & 0.083 \\
            150 & 0.035 & 0.038 & 0.041 \\
            200 & 0.029 & 0.031 & 0.033 \\
            \bottomrule
        \end{tabular}
    \end{minipage}

    \vspace{-12pt}
\end{table*}

\paragraph{Sensitivity to Input Corruption.}
An observed modality may still be weakly informative or corrupted, which is not captured by a binary mask. 
To test whether the fused temperature responds to input quality beyond modality presence, we analyze PAMAP2 in the complete-modality setting with additive Gaussian noise:
\[
\widetilde{x}
=
x+\eta,
\qquad
\eta
\sim
\mathcal{N}\!\left(
0,
(\alpha \sigma_k)^2
\right),
\]
where \(\alpha\in\{0,0.5,0.9\}\) controls the corruption severity and \(\sigma_k\) is the standard deviation of client \(k\)'s training data. As shown in Table~\ref{tab:noisy_modalities}, $\tau_f$ increases monotonically with the corruption level at every evaluated round (e.g., $0.063 \rightarrow 0.075 \rightarrow 0.083$ at round 100). This indicates that $\tau_f$ responds to input quality rather than modality presence alone, assigning higher temperatures to more corrupted inputs.

\begin{wraptable}{r}{0.5\textwidth}
    \centering
    \vspace{-10pt}

    \caption{
        System cost and macro-F1 on PAMAP2.
    }
    \label{tab:cost_analysis}

    \small
    \setlength{\tabcolsep}{1.5pt}
    \renewcommand{\arraystretch}{0.98}

    \resizebox{\linewidth}{!}{%
        \begin{tabular}{@{}lccc@{}}
            \toprule
            \textbf{Method}
            & \textbf{Infer. time $\downarrow$}
            & \textbf{Train GFLOPs $\downarrow$}
            & \textbf{F1 score $\uparrow$} \\
            \midrule
            FedAvg  & 1.14 & 11.40  & 0.634 \\
            FedProx & 1.09 & 11.40  & 0.636 \\
            MOON    & 1.10 & 34.19  & 0.636 \\
            FedPer  & 1.11 & 11.40  & 0.644 \\
            FedRoD  & 1.16 & 13.97  & 0.664 \\
            PmcmFL  & 1.71 & 11.40  & 0.643 \\
            PEPSY   & 1.91 & 109.60 & 0.641 \\
            \midrule
            \textbf{\system{}}
            & 1.44
            & 19.40
            & \textbf{0.683} \\
            \bottomrule
        \end{tabular}%
    }

    \vspace{-8pt}
\end{wraptable}
\paragraph{Cost Analysis.}
\system{} adds lightweight modality-specific confidence heads during training, which are omitted at inference, and a client-private task head. Table~\ref{tab:cost_analysis} compares inference time, training GFLOPs, and macro-F1 on PAMAP2 under the (Mod, Mod) dual-axis missingness regime. \system{} achieves the highest macro-F1 ($0.683$) with $19.40$ GFLOPs per global round. Although this exceeds FedRoD's $13.97$ GFLOPs, it remains well below MOON ($34.19$) and PEPSY ($109.60$). Its inference time is $1.44$ seconds, compared with $1.16$ for FedRoD, $1.71$ for PmcmFL, and $1.91$ for PEPSY. Overall, \system{} incurs moderate additional cost over FedRoD for a $1.9$-point macro-F1 gain, while remaining substantially cheaper to train than MOON and PEPSY.

\begin{wraptable}{r}{0.5\textwidth}
    \centering
    \vspace{-10pt}

    \caption{
        Ablation macro-F1 on PAMAP2.
    }
    \label{tab:ablation}

    \small
    \setlength{\tabcolsep}{2.5pt}
    \renewcommand{\arraystretch}{1.3}

    \resizebox{\linewidth}{!}{%
        \begin{tabular}{@{}lccc@{}}
            \toprule
            \textbf{Variant}
            & \textbf{(High, Mod)}
            & \textbf{(High, High)}
            & \textbf{Average} \\
            \midrule

            Single-path fused conf.
            & 0.594 \scalebox{\std}{± 0.007}
            & 0.468 \scalebox{\std}{± 0.003}
            & 0.531 \scalebox{\std}{± 0.005} \\

            Dual-path fused conf.
            & 0.621 \scalebox{\std}{± 0.012}
            & 0.467 \scalebox{\std}{± 0.012}
            & 0.544 \scalebox{\std}{± 0.012} \\

            Shared+private temp.
            & 0.616 \scalebox{\std}{± 0.017}
            & 0.482 \scalebox{\std}{± 0.007}
            & 0.549 \scalebox{\std}{± 0.012} \\

            w/o fused-repr.\ SG
            & \textbf{0.653 \scalebox{\std}{± 0.003}}
            & 0.497 \scalebox{\std}{± 0.008}
            & 0.575 \scalebox{\std}{± 0.005} \\

            w/o unimodal SG
            & 0.638 \scalebox{\std}{± 0.011}
            & 0.500 \scalebox{\std}{± 0.022}
            & 0.569 \scalebox{\std}{± 0.017} \\

            \midrule

            \system{} (complete)
            & 0.652 \scalebox{\std}{± 0.004}
            & \textbf{0.514 \scalebox{\std}{± 0.017}}
            & \textbf{0.583 \scalebox{\std}{± 0.011}} \\

            \bottomrule
        \end{tabular}%
    }

    \vspace{-8pt}
\end{wraptable}
\paragraph{Ablation Study.}
Table~\ref{tab:ablation} evaluates five ablations of \system{} on PAMAP2 under high inter-client heterogeneity with moderate and high intra-client incompleteness. \emph{Single-path fused conf.} uses one task pathway and a fusion-level confidence; \emph{Dual-path fused conf.} adds shared-private separation; and \emph{Shared+private temp.} introduces modality-specific confidence heads while tempering both pathways. Two additional variants remove stop-gradient at the fused-representation or unimodal-confidence interface.
The complete \system{} achieves the highest average macro-F1 ($0.583$) and the best result under (High, High) ($0.514$). 
Shared-private separation raises the average from $0.531$ to $0.544$, modality-specific confidence further improves it to $0.549$, and private-only tempering increases it to $0.583$, a gain of $0.034$. Removing the fused-representation or unimodal stop-gradient reduces the average to $0.575$ and $0.569$, respectively. Although removing the fused-representation stop-gradient slightly improves (High, Mod) performance ($0.653$ vs.\ $0.652$), it degrades the more incomplete (High, High) setting from $0.514$ to $0.497$. These results support modality-specific confidence estimation, private-only tempering, and stop-gradient isolation at both interfaces.

\section{Conclusion}\label{s:conclusion}
We studied multimodal federated learning under \emph{dual-axis modality missingness}, where modality availability varies across clients and individual samples may contain only subsets of locally available modalities. 
We proposed \system{}, which combines modality-aware confidence tempering with gradient-decoupled private adaptation. The
method fuses per-modality confidence estimates into a sample-adaptive training temperature and confines confidence-tempered updates to a client-private pathway, reducing their influence on shared federated optimization. Across diverse health-sensing and biomedical benchmarks, \system{} achieved the highest average macro-F1 on every dataset. Further analyses show favorable calibration and that the learned temperature responds to modality incompleteness and input corruption, while private-only tempering reduces shared-gradient instability and client-to-global update drift.
\clearpage

\bibliography{ref}
\bibliographystyle{plain}

\newpage

\appendix

\section{Algorithm}
\label{app:algorithm}

\begin{algorithm}[H]
\caption{Training procedure of \system{}}
\label{alg:flux}
\footnotesize
\begin{algorithmic}[1]
\Require Client datasets $\{\mathcal{D}_k\}_{k=1}^{K}$;
shared parameters
$\theta=\{\{\theta_{E_m},\theta_{C_m}\}_{m=1}^{M},
\theta_F,\theta_G\}$;
private parameters $\{\phi_k\}_{k=1}^{K}$;
rounds $T$; local epochs $E$

\For{$t=0,\ldots,T-1$}
    \State Select $\mathcal{S}^{(t)}$ and broadcast $\theta^{(t)}$

    \ForAll{$k\in\mathcal{S}^{(t)}$ \textbf{in parallel}}
        \State $\theta_k\gets\theta^{(t)}$; load retained $\phi_k$

        \For{$e=1,\ldots,E$}
            \ForAll{minibatches $\mathcal{B}\subset\mathcal{D}_k$}
                \State $\mathcal{L}_{\mathrm{shared}}^{\mathcal B}
                \gets 0$;
                $\mathcal{L}_{\mathrm{priv}}^{\mathcal B}\gets 0$

                \ForAll{$(\mathbf{x},y,\mathbf{a})\in\mathcal{B}$}
                    \State $\mathcal{O}\gets\{m:a_m=1\}$

                    \ForAll{$m\in\mathcal{O}$}
                        \State $h_m\gets E_m(x_m)$
                        \State $(z_m,\nu_m)
                        \gets C_m(\operatorname{sg}(h_m))$
                        \State $\tau_m\gets\exp(\nu_m/2)$
                    \EndFor

                    \State $\displaystyle
                    \mathcal{L}_{\mathrm{uni}}
                    \gets
                    \frac{1}{|\mathcal{O}|}
                    \sum_{m\in\mathcal{O}}
                    \operatorname{CE}\!\left(
                    \frac{z_m}{\tau_m},y
                    \right)$

                    \State $\displaystyle
                    \tau_f
                    \gets
                    \left(
                    \sum_{m\in\mathcal{O}}
                    \frac{1}{\tau_m^2+\epsilon}
                    +\epsilon
                    \right)^{-1/2}$

                    \State $\displaystyle
                    h_f
                    \gets
                    F\!\left(
                    \{h_m\}_{m\in\mathcal{O}},
                    \mathbf{a}
                    \right)$
                    \State $z_G\gets G(h_f)$
                    \State $z_{P,k}\gets P_k(\operatorname{sg}(h_f))$

                    \State $\displaystyle
                    \mathcal{L}_{\mathrm{shared}}^{\mathcal B}
                    \gets
                    \mathcal{L}_{\mathrm{shared}}^{\mathcal B}
                    +\operatorname{CE}(z_G,y)
                    +\mathcal{L}_{\mathrm{uni}}$

                    \State $\displaystyle
                    \mathcal{L}_{\mathrm{priv}}^{\mathcal B}
                    \gets
                    \mathcal{L}_{\mathrm{priv}}^{\mathcal B}
                    +\operatorname{CE}\!\left(
                    \frac{z_{P,k}}{\operatorname{sg}(\tau_f)},
                    y
                    \right)$
                \EndFor

                \State $\displaystyle
                \mathcal{L}_{\mathrm{shared}}^{\mathcal B}
                \gets
                \mathcal{L}_{\mathrm{shared}}^{\mathcal B}/|\mathcal{B}|$
                \State $\mathcal{L}_{\mathrm{priv}}^{\mathcal B}
                \gets
                \mathcal{L}_{\mathrm{priv}}^{\mathcal B}/|\mathcal{B}|$

                \State Update $\theta_k$ using
                $\mathcal{L}_{\mathrm{shared}}^{\mathcal B}$
                \State Update $\phi_k$ using
                $\mathcal{L}_{\mathrm{priv}}^{\mathcal B}$
            \EndFor
        \EndFor

        \State Return $\theta_k$; retain updated $\phi_k$ locally
    \EndFor

    \State Set each $\theta_m^{(t+1)}$ by aggregating
    $\{(\theta_{E_m,k},\theta_{C_m,k}) :
    k\in\mathcal{S}^{(t)},\, n_{k,m}>0\}$,
    weighted by $n_{k,m}$
    
    \State Set $(\theta_F^{(t+1)},\theta_G^{(t+1)})$
    by aggregating the corresponding client parameters over
    $\mathcal{S}^{(t)}$, weighted by $n_k$

\EndFor
\end{algorithmic}
\end{algorithm}

\section{Construction of Dual-Axis Modality Missingness Regimes}
\label{app:missingness_simulation}
Existing multimodal federated learning studies often model either
sample-level dropout or fixed client-level modality subsets
~\citep{zhao2022multimodal,bao2023multimodal,feng2023fedmultimodal}.
We instead distinguish two coupled axes: \emph{intra-client modality
incompleteness}, which determines the modalities observed for each sample,
and \emph{inter-client modality heterogeneity}, which determines the modalities
available to each client. We instantiate these axes according to the data
source. For the three health-sensing datasets, both axes are simulated. For
ADNI, we retain naturally missing patient modalities and vary only
client-level modality availability.

\noindent\textbf{Health-Sensing Datasets.}
Health-sensing deployments may exhibit both static differences in device
configurations and temporally correlated sensor failures~\cite{opportunity}.
We model client-level modality availability using a Beta--Bernoulli
process.
In particular, let \(k\in\{1,\ldots,K\}\) index clients,
\(m\in\{1,\ldots,M\}\) index sensing modalities, and
\(i\in\{1,\ldots,n_k\}\) index the temporally ordered samples
of client \(k\). Here, each modality corresponds to a sensor stream or
sensor-location pair defined for the respective dataset.
 Specifically, each client \(k\) draws an availability probability
\(q_k\), followed by a binary availability indicator for each modality:
\[
q_k \sim \operatorname{Beta}(\alpha_a,\beta_a),
\qquad
c_{k,m}\mid q_k
\sim
\operatorname{Bernoulli}(q_k),
\]
where \(c_{k,m}=1\) indicates that modality \(m\) belongs to the static
sensor suite of client \(k\). Conditional on \(q_k\), the indicators
\(\{c_{k,m}\}_{m=1}^{M}\) are sampled independently. In the
\emph{homogeneous} inter-client setting, we set \(c_{k,m}=1\) for all
clients and modalities. The \emph{moderate} and \emph{high}
inter-client heterogeneity settings use
\[
(\alpha_a,\beta_a)=(45,20)
\quad\text{and}\quad
(\alpha_a,\beta_a)=(45,45),
\]
respectively. These configurations correspond to expected modality
availability rates of \(45/65\approx0.692\) and \(45/90=0.5\).
In the homogeneous setting, all clients have the same set of sensor modalities, although sample-level sensor failures may still occur.
For each modality available at a client, i.e., \(c_{k,m}=1\), we model
its operational state across temporally ordered samples using a
two-state Markov chain:
\[
s_{k,i,m}\in\{0,1\},
\qquad
\mathbf{P}
=
\begin{bmatrix}
p_{00} & p_{01}\\
p_{10} & p_{11}
\end{bmatrix},
\]
where
\[
p_{uv}
=
\Pr\!\left(
s_{k,i,m}=v
\mid
s_{k,i-1,m}=u
\right),
\qquad u,v\in\{0,1\}.
\]
State \(0\) denotes that the modality is missing, whereas state \(1\)
denotes that it is observed. A large \(p_{00}\) produces persistent
missing segments, while a large \(p_{11}\) produces persistent observed
segments, yielding temporally structured failures rather than
independent sample-wise dropout.
The resulting sample-level observation mask is
\[
a_{k,i,m}=c_{k,m}s_{k,i,m}.
\]

Modality \(m\) is observed for sample \(i\) at client \(k\) only
when it belongs to the client's static modality set and is operational
for that particular sample.
The client-level variable \(c_{k,m}\) therefore defines
inter-client modality heterogeneity, whereas \(s_{k,i,m}\) defines
temporally varying intra-client modality incompleteness.

\begin{figure*}[t!]
    \centering
    \includegraphics[width=.9\textwidth]{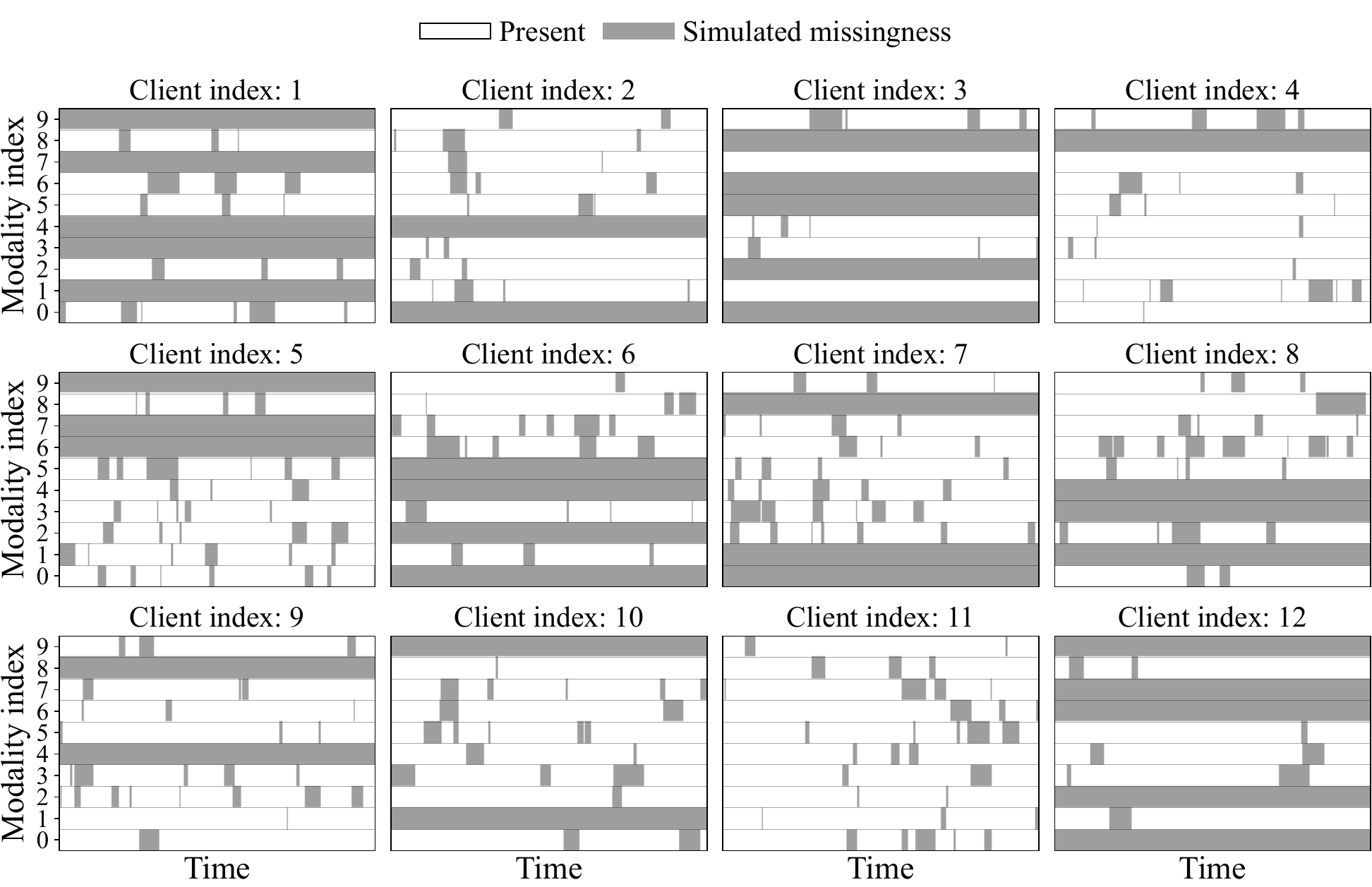}
    \caption{Missingness patterns for 12 RealWorldHAR clients under moderate
    inter-client heterogeneity and moderate intra-client incompleteness.
    Clients have different static sensor suites, while available sensors
    exhibit temporally bursty failures.}
    \label{fig:inter_client_missing}
\end{figure*}
Figure~\ref{fig:inter_client_missing} illustrates the resulting dual-axis
patterns. Unlike i.i.d. dynamic dropout~\citep{feng2023fedmultimodal} or
purely static modality removal
~\citep{bao2023multimodal,zhao2022multimodal}, this construction captures
both population-level differences in sensor ownership and within-client
temporal sensor instability.

\paragraph{ADNI Dataset.}
Let \(k\) index silos, which serve as FL clients; \(i\) index patients
assigned to silo \(k\); and \(m\in\{1,\ldots,4\}\) index structural MRI,
genomic profiles, clinical assessments, and biospecimen
measurements. Let \(r_{k,i,m}\in\{0,1\}\) denote the naturally occurring
observation mask. We construct a five-silo partition using multilabel
stratification over diagnosis labels and natural modality masks, producing
similar label and missingness-pattern distributions across silos.
In the \emph{homogeneous} setting, no modality is removed at the silo level:
\[
c_{k,m}=1,
\qquad
a_{k,i,m}=r_{k,i,m}.
\]

For the \emph{moderate} and \emph{high} inter-client heterogeneity settings,
we retain the same patient-to-silo assignment and sample a static modality
profile for each silo:
\[
p_{a,k}\sim\operatorname{Beta}(\alpha_a,\beta_a),
\qquad
c_{k,m}\mid p_{a,k}\sim\operatorname{Bernoulli}(p_{a,k}),
\]
\[
a_{k,i,m}=r_{k,i,m}c_{k,m}.
\]

We use \((\alpha_a,\beta_a)=(17,3)\) for moderate heterogeneity and \((6,14)\) for high heterogeneity. If a silo's sampled modality-availability vector is all-zero, we set one randomly chosen modality to available so each silo retains at least one modality. 
Patients with \(\sum_{m=1}^{M} a_{k,i,m}=0\) after applying simulated missingness to the natural observation mask
are excluded. The \emph{extreme} setting is constructed without stochastic modality
removal. Instead, patients are grouped by their exact nonempty natural
modality mask, such that each client has a unique modality combination.

\section{Related Work}\label{app:related_work}
Prior multimodal FL methods address different missingness regimes. PmcmFL uses prototype masking and contrastive learning for within-client missing modalities, while MFCPL applies complete prototypes and cross-modal regularization under severe missingness
~\citep{bao2023multimodal,le2025cross}. FedMAC handles complete and partial modality missingness through imputation embeddings, and contrastive regularization~\citep{nguyen2024fedmac}. Harmony and FedMEMA focus on heterogeneous modality sets across clients using modality-wise training or modality-specific encoders and multimodal anchors~\citep{ouyang2023harmony,dai2024federated}, while PEPSY addresses cross-client and within-client missing-data patterns through representation reconfiguration~\citep{nguyenlearning}. These methods primarily compensate for missing representations or coordinate learning
across modality configurations.
Personalized FL motivates shared-private parameterization~\citep{arivazhagan2019federated,chen2022on}, while input-dependent uncertainty modeling and adaptive temperature scaling support evidence-dependent confidence control~\citep{kendall2017uncertainties,joy2023sample}. In contrast, \system{} models the reliability of each observed modality during federated training and fuses these estimates into a sample-adaptive temperature that reflects both evidence completeness and quality. It then confines confidence-tempered adaptation to a gradient-decoupled private pathway, preventing evidence-dependent tempering from modifying shared task gradients. Thus, \system{} addresses the optimization tension between sample-specific evidence adaptation and stable cross-client representation learning.

\section{Experiment Details}\label{app:experiment_details}
\begin{table*}[t]
\centering
\caption{Hyperparameter configurations.}
\label{tab:exp_hyperparameters}
\begin{tabular}{@{}llcccc@{}}
\toprule
\textbf{Method} & \textbf{HP} & \textbf{PAMAP2} & \textbf{RealWorldHAR} & \textbf{SleepEDF} & \textbf{ADNI} \\
\midrule
\midrule
FedAvg & Learning Rate & 0.001 & 0.03 & 0.03 & 0.0001 \\
\midrule
FedPer & Learning Rate & 0.001 & 0.03 & 0.001 & 0.0001 \\
\midrule
\multirow{2}{*}{FedProx} & Learning Rate & 0.001 & 0.03 & 0.01 & 0.0001 \\
 & Proximal Term & 0.1 & 0.01 & 0.01 & 0.01 \\
\midrule
\multirow{3}{*}{MOON} & Learning Rate & 0.001 & 0.03 & 0.03 & 0.0001 \\
 & Contrast. Weight & 10 & 0.1 & 10 & 0.1 \\
 & Temperature & 0.5 & 0.5 & 1.0 & 1.0 \\
\midrule
\multirow{2}{*}{PmcmFL} & Learning Rate & 0.001 & 0.03 & 0.001 & 0.0001 \\
 & CLIP Loss Weight & 0.1 & 0.01 & 0.5 & 0.5 \\
\midrule
FedRoD & Learning Rate & 0.001 & 0.03 & 0.01 & 0.0001 \\
\midrule
PEPSY & Learning Rate & 0.01 & 0.01 & 0.001 & 0.00004 \\
\midrule
\multirow{2}{*}{\system{}} & Learning Rate & 0.001 & 0.001 & 0.001 & 0.0001 \\
 & Conf.\ Head Hidden Dim. & 512 & 512 & 512 & 128 \\
\bottomrule
\end{tabular}
\end{table*}

\subsection{Datasets}\label{app:dataset_description}
We use four real-world multimodal datasets: three health-sensing benchmarks (PAMAP2~\citep{reiss2012introducing}, RealWorldHAR~\citep{sztyler2016body}, and SleepEDF~\citep{goldberger2000physiobank, kemp2000analysis}) and the biomedical cohort ADNI~\citep{weiner2010alzheimer}, which contains naturally missing data.

\noindent\textbf{PAMAP2}~\citep{reiss2012introducing} consists of recordings from nine users performing twelve activities using wearable Inertial Measurement Unit (IMU) sensors. Following prior work~\citep{jain2022collossl}, we exclude one subject who contributed data for only a single activity, resulting in eight clients. The dataset provides accelerometer and gyroscope signals from three body locations: wrist, chest, and ankle, yielding six distinct sensing modalities in total.

\noindent\textbf{SleepEDF}~\citep{goldberger2000physiobank, kemp2000analysis} contains sleep recordings from 20 participants, including electroencephalography~(EEG), electrooculography~(EOG), chin electromyography~(EMG), respiration signals, and event markers. Each recording is annotated with hypnograms containing five sleep stages. Following prior work~\citep{tsinalis2016automatic, phan2018joint}, we utilize the Sleep Cassette subset, which focuses on age-related sleep patterns in healthy individuals and is commonly used for sleep-stage classification.

\noindent\textbf{RealWorldHAR}~\citep{sztyler2016body} consists of activity recordings from fifteen participants performing eight daily activities. Data were collected with seven body-worn IMU sensors, two of which were discarded due to insufficient activity coverage. The final dataset comprises signals from ten modalities, spanning five body locations and two IMU sensor types.

\noindent\textbf{ADNI}~\citep{weiner2010alzheimer} is a multimodal study of Alzheimer's disease and aging. Each patient may contribute up to four data types: structural MRI, genomic profiles, clinical assessments, and biospecimen measurements.
We cast the task as three-way diagnostic classification among cognitively normal (CN), mild cognitive impairment (MCI), and Alzheimer's disease (AD). Following our federated evaluation protocol, participants are partitioned into five silos that serve as FL clients. Unlike the wearable benchmarks, modality incompleteness in ADNI is \emph{natural}, meaning that not all patients have all four modalities available. 
In our experiments, we retain the naturally occurring intra-client modality incompleteness and vary only inter-client modality heterogeneity
by changing modality availability across silos.

\subsection{Baselines}\label{app:baseline_description}
\noindent\textbf{FedAvg}~\citep{mcmahan2017communication} is the standard federated learning baseline that enables collaborative training without sharing raw data but provides no explicit mechanism for handling missing modalities.

\noindent\textbf{FedProx}~\citep{li2020federatedprox} addresses system and statistical heterogeneity. It enhances performance by adding a proximal term to the local training loss, penalizing deviations between local and global models to improve stability and convergence. 

\noindent\textbf{MOON}~\citep{li2021model} targets the problem of local data heterogeneity. It incorporates contrastive learning into federated learning, encouraging alignment between the global and local models’ embeddings while pushing apart embeddings from the client’s previous local model.

\noindent\textbf{FedPer}~\citep{arivazhagan2019federated} addresses statistical heterogeneity by splitting models into shared base layers and client-specific personalization layers. The base layers are trained collaboratively across clients using FedAvg, while the personalization layers are updated only with local data. 

\noindent\textbf{FedRoD}~\citep{chen2022on} bridges generic and personalized federated learning through a decoupled prediction architecture. It decouples the local model into two predictors: a generic head trained with balanced risk minimization to improve robustness against non-IID class distributions, and a personalized head trained with empirical risk minimization to capture client-specific patterns. 

\noindent\textbf{PmcmFL}~\citep{bao2023multimodal} introduces a prototype library to address the challenges of missing modalities in federated multimodal learning. Prototypes are used both as masks for absent modalities and as anchors in a contrastive loss to reduce client heterogeneity.

\noindent\textbf{PEPSY}~\citep{nguyenlearning} tackles the multimodal FL regime where clients have different modality subsets and also suffer within-modality missing features. It learns client-side data-missing profiles to adapt and align the shared representation to each client’s missingness pattern, and aggregates these signals across clients with similar patterns.

\subsection{Training and Evaluation Details}\label{app:learning_setup_details}

\paragraph{Hyperparameter tuning.}
Table~\ref{tab:exp_hyperparameters} reports the final selected
hyperparameters for each method and dataset. For the sensing datasets, we
tune the learning rate over
\(\{0.001,0.01,0.03,0.05\}\) and select the value with the best validation
performance. For ADNI, we use an ADNI-specific grid consisting of local
learning rates in \(\{4\times10^{-5},10^{-4}\}\) and batch sizes in
\(\{8,16\}\), due to its larger input dimensionality and memory footprint.

We additionally tune the FedProx proximal coefficient
\(\mu_{\mathrm{prox}}\) over
\(\{0.001,0.01,0.1,0.5,1\}\) for the sensing datasets and
\(\{0.01,0.1\}\) for ADNI. For MOON, we tune the contrastive weight
\(\mu_{\mathrm{contrast}}\) over \(\{0.1,1,5,10\}\) and the temperature
\(\tau\) over \(\{0.1,0.5,1\}\) for the sensing datasets; for ADNI, the
corresponding grids are \(\{0.1,10\}\) and \(\{0.5,1.0\}\). For PmcmFL,
we tune the CLIP loss weight over
\(\{0.01,0.1,0.5,1.0,5.0\}\) for the sensing datasets and
\(\{0.01,0.1,0.5\}\) for ADNI. For \system{}, we tune the confidence-head hidden dimension over
\(\{128,256,512\}\) on all datasets.

\paragraph{Model selection.}
For global methods (FedAvg, FedProx, MOON, PmcmFL, and PEPSY) the
server selects the checkpoint with the highest mean validation macro-F1
across clients. For personalized methods (FedPer, FedRoD, and
\system{}) each client independently selects the checkpoint with the
highest local validation macro-F1. Final test performance is computed
using the selected global or client-specific checkpoints, respectively.

\paragraph{Evaluation consistency and hardware.}
For each dataset, missingness regime, and random seed, all methods used
the same train, validation, and test partitions and the same generated
client- and sample-level modality masks. Experiments were conducted using
Intel Xeon CPUs and NVIDIA RTX 3090 GPUs. The server was equipped with
432~GB of RAM and ran Ubuntu 22.04.2 LTS with CUDA 11.5.

\end{document}